\documentclass[conference]{IEEEtran}
\IEEEoverridecommandlockouts
\usepackage{cite}
\usepackage{amsmath,amssymb,amsfonts}
\usepackage{algorithmic}
\usepackage{textcomp}
\usepackage{graphicx}
\usepackage{xcolor}
\usepackage{multirow}
\usepackage{subfigure}
\usepackage{hyperref}

\def\BibTeX{{\rm B\kern-.05em{\sc i\kern-.025em b}\kern-.08em
    T\kern-.1667em\lower.7ex\hbox{E}\kern-.125emX}}

\begin{document}

\title{MTSMAE: Masked Autoencoders for Multivariate Time-Series Forecasting
}

\author{
	\IEEEauthorblockN{Peiwang Tang$^{1, 2}$, Xianchao Zhang$^{3, 4*}$}
	\IEEEauthorblockA{$^{1}$ Institute of Advanced Technology, University of Science and Technology of China, China}
	\IEEEauthorblockA{$^{2}$ G60 STI Valley Industry \& Innovation Institute, Jiaxing University, China}
	\IEEEauthorblockA{$^{3}$ Key Laboratory of Medical Electronics and Digital Health of Zhejiang Province,
		\\ Jiaxing University, China}
	\IEEEauthorblockA{$^{4}$ Engineering Research Center of Intelligent Human Health Situation Awareness of Zhejiang Province,
		\\ Jiaxing University, China}
	\IEEEauthorblockA{\{tpw\}@mail.ustc.edu.cn, \{zhangxianchao\}@zjxu.edu.cn}
}

\maketitle

\begin{abstract}
Large-scale self-supervised pre-training Transformer architecture have significantly boosted the performance for various tasks in natural language processing (NLP) and computer vision (CV).
However, there is a lack of researches on processing multivariate time-series by pre-trained Transformer, and especially, current study on masking time-series for self-supervised learning is still a gap.
Different from language and image processing, the information density of time-series increases the difficulty of research. The challenge goes further with the invalidity of the previous patch embedding and mask methods.
In this paper, according to the data characteristics of multivariate time-series, a patch embedding method is proposed, and we present an self-supervised pre-training approach based on Masked Autoencoders (MAE), called MTSMAE, which can improve the performance significantly over supervised learning without pre-training. 
Evaluating our method on several common multivariate time-series datasets from different fields and with different characteristics, experiment results demonstrate that the performance of our method is significantly better than the best method currently available.
\end{abstract}

\begin{IEEEkeywords}
Autoencoder, Pre-Training, Time-Series, Forecasting
\end{IEEEkeywords}

\begin{figure*}[th]
	\centering
	\subfigure[Pre-Training]{
		\includegraphics[scale=0.9]{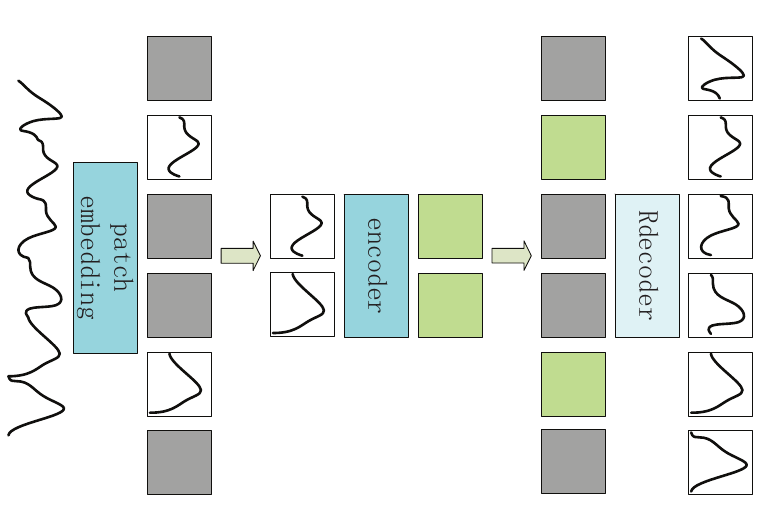}
		\label{fig1}
	}
	\subfigure[Fine-Tuning]{
		\includegraphics[scale=0.9]{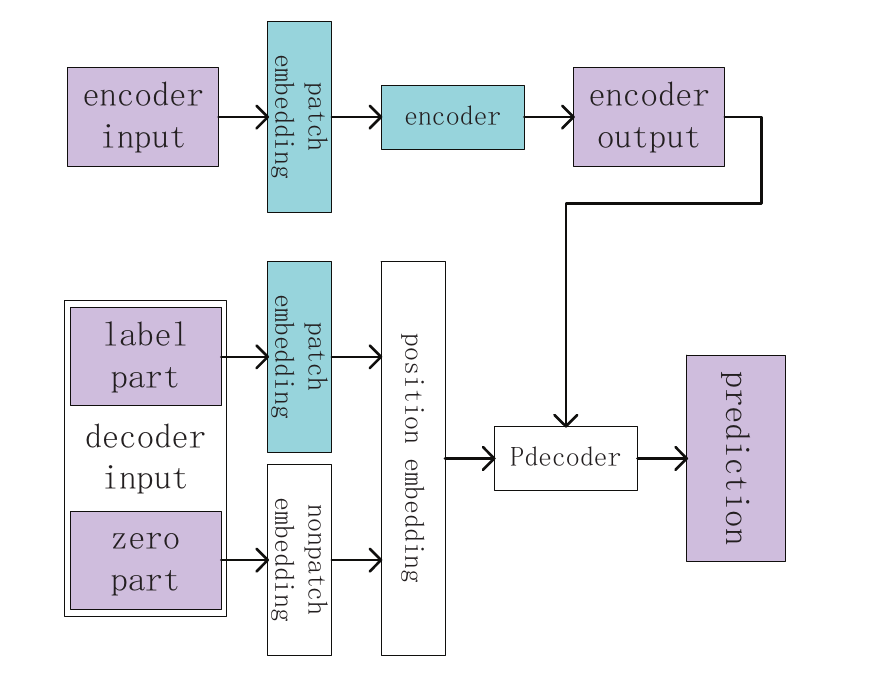}
		\label{fig2}
	}
	\caption{Our MTSMAE architecture. In the pre-training, our model consists of encoder and Rdeocder (the decoder responsible for recovering the original input); in the fine-tuning, our model consists of encoder and Pdeocder (the decoder responsible for predicting future data).}
	\label{fig0}
\end{figure*}

\section{Introduction}
With the fast evolvement of deep learning in recent years \cite{krizhevsky2012imagenet, vaswani2017attention, tang2022features}, training a model is anticipated to accommodate hundreds of millions of labeled data \cite{2021An}. The demand for large scale data processing has been solved by self-supervised pre-training in natural language processing (NLP) and computer vision (CV) fields \cite{devlin2018bert, he2022masked}. Most of these solutions are based on masked modeling, such as masked language modeling in NLP \cite{radford2018improving, radford2019language} or masked image modeling in CV \cite{chen2020generative, bao2021beit ,xie2022simmim}. Their ideas are conceptually simple: firstly mask parts of the data based on the original data and then enable to recovery these parts by learning \cite{geng2022multimodal, li2022uniform}. Masked modeling encourages the model to infer the deleted parts according to the context information, so that the model can learn the deep semantics, which has become the benchmark of self-supervised pre-training in NLP and CV fields \cite{devlin2018bert, he2022masked}. These pre-trained masked modeling has been proved to be well applied to various downstream tasks, among which a simpler and more effective way is masked autoencoders (MAE) \cite{he2022masked}. However, despite a widely interests in this idea from academia and industry following the success of MAE, the progress of autoencoder methods in the field of multivariate time-series data (MTSD) lags behind other fields. 

One of the main reasons is that the information density of MTSD is different from that of CV and NLP. The local information of MTSD seems to be heavy spatial redundancy, but the multivariate information within each time point has high specificity. Missing information can be easily learned from information at adjacent time points with little high-level understanding. In order to overcome this difference and encourage learning more useful features, we use the idea of Vision Transformer (ViT) \cite{2021An}, patch MTSD, and mask more random patches than the original MAE, e.g. 85$\%$. This simple strategy has a ideal performance in MSTD, which can reduce redundancy effectively and futher increase the overall understanding beyond low-level information of the model.

Another reason is the design of the decoder. The decoder of the autoencoder maps the latent representation back to the input. In CV, the decoder can  reconstruct the pixel level representation in patch. In NLP, the decoder predicts missing words. In MTSD, the decoder recovers completely different data with information specificity and a dimension that even highly to 321. For different types of MTSD, the dimensions of data at each time point are different, which may have 321 dimensions or only 7 dimensions. We find that the design of decoder plays a key role in the latent representation of learning for MTSD.

Based on the above analysis, we propose a very simple and effective MTSMAE for MTSD representation learning. Our MTSMAE idea is simple: In the pre-training, patch MTSD, masks random patches from the input and recover the missing patches; in the fine-tuning, take out the encoder trained in the previous step, and the input of the decoder is redesigned. In addition, our encoder only calculates visible patches. Unlike the decoder with only one layer of MLP in Bert \cite{devlin2018bert}, we design different levels of decoder according to different MTSD, but compared with MAE, our decoder are all lightweight. We have conducted extensive experiments on four different datasets of three types. The final experimental results show that our proposed MTSMAE can significantly improve the accuracy of prediction, and is superior to other state-of-the-art models.

\section{Related work}

\subsection{Masked modeling for Self-supervised learning}
Self-supervised learning approaches have aroused great interest in natural language processing and computer vision, often focusing on different pretext tasks for pre-training  \cite{doersch2015unsupervised, wang2015unsupervised, noroozi2016unsupervised}.
BERT \cite{devlin2018bert} and GPT \cite{radford2018improving, radford2019language, brown2020language} are very successful masked modeling for pre-training in NLP. They learn representations from the original input corrupted by masking, which have been proved by a large amount of evidence to be highly extensible \cite{brown2020language}, and these pre-trained representations can be well extended to various downstream tasks.
As the original method applied BERT to CV, BEIT \cite{bao2021beit} firstly ``tokenize" the original image as a visual token and then randomly mask some image patches and input them into the backbone Transformer \cite{vaswani2017attention}.
The goal of pre-training is to restore the original visual token from the damaged image patch. 
Based masked modeling as all autoencoder \cite{hinton1993autoencoders}, MAE \cite{he2022masked} uses the encoder to map the observed signal to the potential representation, and the decoder to reconstruct the original signal from the latent representation. In turn, different from the classic autoencoder, MAE adopts an asymmetric design to allow the encoder to operate on only part of the observed signal (without mask token) and use a lightweight decoder to reconstruct the complete signal from the latent representation along with mask tokens.

\subsection{Models for Time-Series Forecasting}
Forecasting is one of the most important applications of time-series.
LSTNet \cite{lai2018modeling} uses convolutional neural network (CNN) \cite{krizhevsky2012imagenet} and recurrent neural network (RNN) to extract short-term local dependence patterns between variables and find long-term patterns of time-series trends. In addition, the traditional autoregressive model is used to solve the scale insensitivity  problem of neural network model.
The temporal convolutional network (TCN) \cite{bai2018empirical} is proposed to make the CNN have time-series characteristics, as a variant of CNN that deals with sequence modeling tasks. It mixes RNN and CNN architecture to use the causal convolution to simulate temporal causality.
Reformer \cite{kitaev2019reformer} is proposed to replace the original dot-product attention with a new one using locality-sensitive hashing. It decreases the complexity from $ \mathcal{O}  (L_{2} )$ to $ \mathcal{O}  (L \log_{}{L} )$ and makes the storage activated only once in the training process rather than $n$ times (here $n$ refers to the number of layers) by using reversible residual layers to replace standard residuals.
LogTrans \cite{li2019enhancing} proposes a convolutional self-attention mechanism, which uses causal convolution to process the local context of the sequence and calculate the query / key of self-attention. Further more, it uses the logsparse Transformer architecture to ensure that each cell can receive signals from other cells in the sequence data, while reducing the time complexity of the architecture. 
Informer \cite{zhou2021informer} is proposed ProbSparse self-attention and encoder with self-attention distilling. The former is based on query and key similarity sampling dot-product pairs, which reduces the computational complexity of Transformer and allows it to accept longer input. The latter adopts the concept of distillation to design encoder, so that the model can continuously extract the feature vectors that focus on, while reducing the memory occupation.

\begin{figure*}[t]
	\centering
	\includegraphics[scale=0.9]{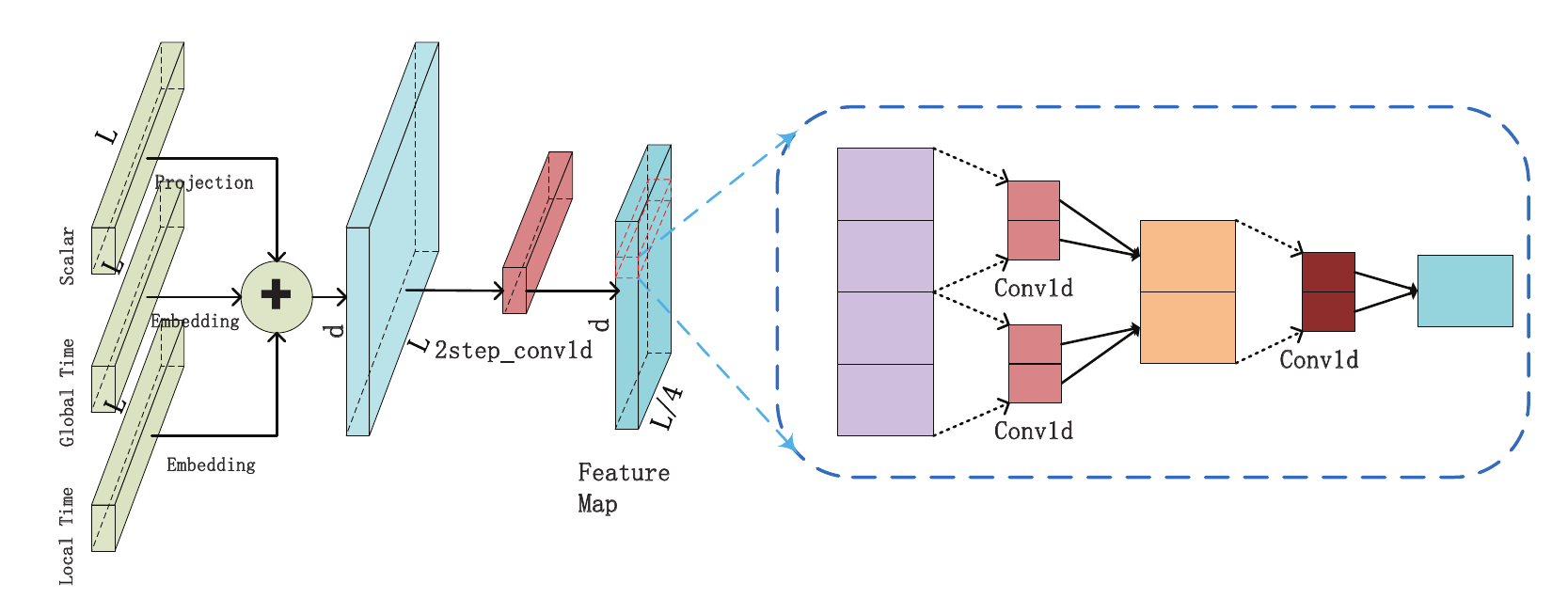}
	\caption{Patch embedding method.}
	\label{img1}
\end{figure*}

\section{Methodology}

The problem of multivariate time-series forecasting is to input the past sequence $\mathcal{X}^{t}=\left \{x^t_{1}, \cdots ,x^t_{L_x} |x^t_{i}\in \mathbb{R}^{d_x} \right \}$ at time $t$, and output the predict the corresponding future sequence $\mathcal{Y}^{t}=\left \{y^t_{1}, \cdots ,y^t_{L_y} |y^t_{i}\in \mathbb{R}^{d_y} \right \}$, where $ L_x$ and $ L_y$ are the lengths of input and output sequences respectively, and $d_x$ and $d_y$ are the feature dimensions of input $\mathcal{X}$ and output $\mathcal{Y}$ respectively.

Our masked autoencoders (MTSMAE) is a simple autoencoding method and the training process is divided into two stages, as shown in the Fig.~\ref{fig0}. As all autoencoders, there is an encoder and a decoder in our method. The encoder maps the observed signal to a latent representation and the decoder reconstructs the original signal from the latent representation in the pre-training, or output $\mathcal{Y}$ in the fine-tuning. In the pre-training, the encoder only operates the partial signal (without mask token) observed.

\subsection{Patch embedding}\label{AA}

As shown in the Fig.~\ref{img1}, the input embedding consists of three parts, a scalar projection SP (we project the scalar context $x^t_i $ into $ d_{model}$-dim vector with 1-D convolutional filters. The kernel width of one-dimensional convolutional filter is 3, and stride is 1), a local position embedding PE \cite{vaswani2017attention} and a global time stamp embedding SE \cite{zhou2021informer}:
\begin{equation}\mathcal{X} = SP + PE + SE \end{equation}
which
\begin{equation}SP = Conv1d(x^t_i)\end{equation}
\begin{equation}PE_{\left ( i, 2j \right )} = sin\left ( i / 10000^{2j/d_{model}}  \right )\label{eq1}
\end{equation}
\begin{equation}PE_{\left ( i, 2j+1 \right )} = cos\left ( i / 10000^{2j/d_{model}}  \right )\label{eq2}
\end{equation}
\begin{equation}SE = E(month) + E(day) + E(hour)+ E(minute) 
\end{equation}
where $ i\in \left \{ 1,\dots,L_x  \right \} $, $ j\in \left \{ 1,\dots,\left \lfloor d_{model}/2 \right \rfloor  \right \} $, $d_{model}$ is the feature dimension after embedding, $E$ is a learnable stamp embeddings with limited vocab size (up to 60, namely taking minutes as the finest granularity).

The original MAE continues the idea of ViT, processing image data $\mathcal{X}\in\mathbb{R}^{H \times W\times C} $, where $(H, W)$ is the resolution of the original image, $C$ is the number of channels. For MTSD,  $\mathcal{X} \in \mathbb{R}^{L_x \times d_x} $, the original patch embedding method is no longer applicable. Therefore, unlike the method of patch image data in ViT, we patch MTSD in the direction of time after embedding:
\begin{equation}\mathcal{X}_{h}=Conv1d(\mathcal{X})\in \mathbb{R}^{L_x/P \times d_{model}}
\end{equation}
\begin{equation}\mathcal{X}_{pte}=Conv1d(\mathcal{X}_{h})\in \mathbb{R}^{L_x/P^2 \times d_{model}}
\end{equation}
where the kernel width of one-dimensional convolutional filter and $stride=P$, $ \mathcal{X}_{pte}$ is the final result of the patch embedding. We use two-step one-dimensional convolution to control the resolution of each MSTD patch to $(P, d_{model})$, so the length of the last input sequence $L=L_x/P^2$.

\subsection{Model inputs}

In the pre-training, the input of the encoder is the output $\mathcal{X}_{pte}$ of the past sequence $\mathcal{X}^{t}$ after patch embedding. However, only unmasked patches are input. A random masking method is adopt, that is, the patches are randomly sampled without replacement, and follow the uniform distribution. 
The random sampling can tremendously remove the information redundancy of MSTD by deleting a large number of patches (i.e., high masking rate). Thus, a task that can not easily recover information from the visible nearby patches is created.
Finally, the highly sparse input provides an opportunity to train an efficient encoder. In the pre-training, the decoder only inputs the output of the encoder. 

In the fine-tuning, the input of the encoder is all the output $\mathcal{X}_{pte}$ of the past sequence $\mathcal{X}^{t}$ after patch embedding. As shown in the Fig.~\ref{fig2}, the input of the decoder includes not only the output of the decoder, but also the label part $\mathcal{X}^{t}_{label}$ and the forecast part $\mathcal{X}_{0}$, $nonpatch$ $embedding$ is a common embedding without convolution kernel processing, such as the embedding in Informer. It can be formulated as follows:
\begin{equation}\mathcal{X}_{eni} = Patch(\mathcal{X}^{t})\end{equation}
\begin{equation}\mathcal{X}_{dei} = (\mathcal{X}_{eno},Concat(Patch(\mathcal{X}^{t}_{label}),NonPatch(\mathcal{X}_{0})))\end{equation}
where $\mathcal{X}_{0}$ denotes the placeholder filled with zero, $\mathcal{X}_{eno}$ indicates the output of the encoder, $Patch$ refers to patch embedding and $NonPatch$ refers to $nonpatch$ $embedding$.
\subsection{Encoder and Decoder}
Our encoder is the encoder of transformer. In the pre-training, our encoder embeds only visible, unmasked patches through patch embedding, and then processes the output data through a series of transformer encoder blocks. Our encoder only operates on a small part of the whole set, e.g., only 15$\%$, which can greatly reduce the redundancy of information and increase the overall understanding of the model beyond low-level information. In the fine-tuning, our encoder can see all the patches. Except for this, it is no different from the encoder in the pre-training. The Transformer encoder layers are composed of two sub-blocks. The first is a multi-head self-attention mechanism (MSA), and the second is a simple, position-wise fully connected feed-forward network (MLP). Residual connections \cite{he2016deep} are used around each of the two sub-blocks , and layer normalization (LN) \cite{ba2016layer} is then performed.
\begin{equation}\textbf{z}^{\prime} _{n} = LN(\textbf{z}_{n-1} + MSA(\textbf{z}_{n-1}))\end{equation}
\begin{equation}\textbf{z}_{n} = LN(\textbf{z}^{\prime}_{n} + MLP(\textbf{z}^{\prime}_{n}))\end{equation}
where  $n = 1 \dots N $, $z_0 =\mathcal{X}_{eni} $, $ \mathcal{X}_{eno} = z_n$, $N$ is the number of layers of the encoder, and MLP consists of two linear transformations with a ReLU activation in between.
\begin{equation}MLP(x) = max(0,xW_1 + b_1)W_2 + b_2\end{equation}
$W_1$, $W_2$, $b_1$, $b_2$ are all parameters to be learned, and the parameters of each MLP are different.

In the pre-training, we set up the decoder as MAE. The input to the decoder is a complete set including the visible patches output by encoder and mask tokens, where as the vector of learning, masked tokens are the data to be recovered. Except different inputs, there is no difference between the decoder and encoder in the pre-training. In the fine-tuning, our decoder is also the decoder of the transformer. In addition to the two sub-blocks in each encoder layer, the decoder also inserts a third sub-block in the two sub-blocks, which performs multi-head attention on the output of the encoder layers. In order to prevent positions from attending to subsequent positions, MSA is modified to masked multi-head self-attention (MMSA). This masking ensures that the prediction of position $i$ can only rely on the known outputs of positions less than $i$.
\begin{equation}\textbf{z}^{\prime}_{n} = LN(\textbf{z}_{n-1} + MMSA(\textbf{z}_{n-1}))\end{equation}
\begin{equation}\textbf{z}^{\prime \prime} _{n} = LN(\textbf{z}^{\prime}_{n} + MSA(\textbf{z}^{\prime}_{n},\mathcal{X}_{eno}))\end{equation}
\begin{equation}\textbf{z}_{n} = LN(\textbf{z}^{\prime \prime}_{n} + MLP(\textbf{z}^{\prime \prime}_{n}))\end{equation}

 In the pre-training, our MTSMAE reconstructs the input by recovering the specific value of each masking patch. Each element output by the decoder is a vector that can represent a patch. The last layer of the decoder is a linear projection, whose output channel is $P \times D$, $P$ is the length of the patch, and $D$ is the dimension of the time-series. In the fine-tuning, each element output by the decoder represents the data $y^t_{i}$, and the output channel of the last layer of linear projection is $D$.

Our loss function is calculated by the mean square error (MSE) between the model output data $y_{o}$ (recovery, prediction) and the real data $y$. But in the pre-training, we only calculate the loss on the masking patch, similar to Bert, MAE. 
\begin{equation}\textbf{Loss}(y_{o},y)=\frac{1}{L}\sum_{1\leq i\leq L}(y\widehat{}_i-y_i)^2 \end{equation}
Where $L$ represents the number of masking patch or the length of prediction sequence.
\section{Experiment}

\subsection{Datasets}\label{SCM}
We conducted extensive experiments on four datasets of three types to evaluate the progressiveness of the proposed MTSMAE.
\subsubsection{ETT(Electricity Transformer Temperature)}
The ETT is a key indicator for long-term deployment of the electric power, which collects data from two counties in China from July 2016 to July 2018 for a total of two years. ETT is divided into two types, and each of which has two different datasets including 1-hour-level \{ETTh1, ETTh2\} and 15-minute-level \{ETTm1, ETTm2\}. The data at each time point is composed of the target value ``oil temperature" and six power load features. The size of training set, validation set and test set are 12/4/4 months respectively. We choose ETTm1 and ETTh2 as the experimental dataset.
\subsubsection{ECL (Electricity Consuming Load)}
It collects the electricity consumption (Kwh) of 321 customers, that is, each time point includes the respective electricity consumption of them. Due to the lack of data \cite{li2019enhancing}, we follow the settings in Informer \cite{zhou2021informer} to convert the dataset to hourly consumption for 2 years. The size of training set, validation set and test set are 15/3/4 months respectively.
\subsubsection{WTH(Weather)}
This dataset contains the local climatologicale data of nearly 1600 locations in the United States. The data are collected by once an hour from 2010 to 2013. The data at each time point includes the target value ``wet bulb" and 11 climate features. The size of training set, validation set and test set are 28/10/10 months respectively.
\subsection{Experimental Details}

\begin{figure*}[th]
	\centering
	\subfigure[MTSMAE]{
		\includegraphics[scale=0.33]{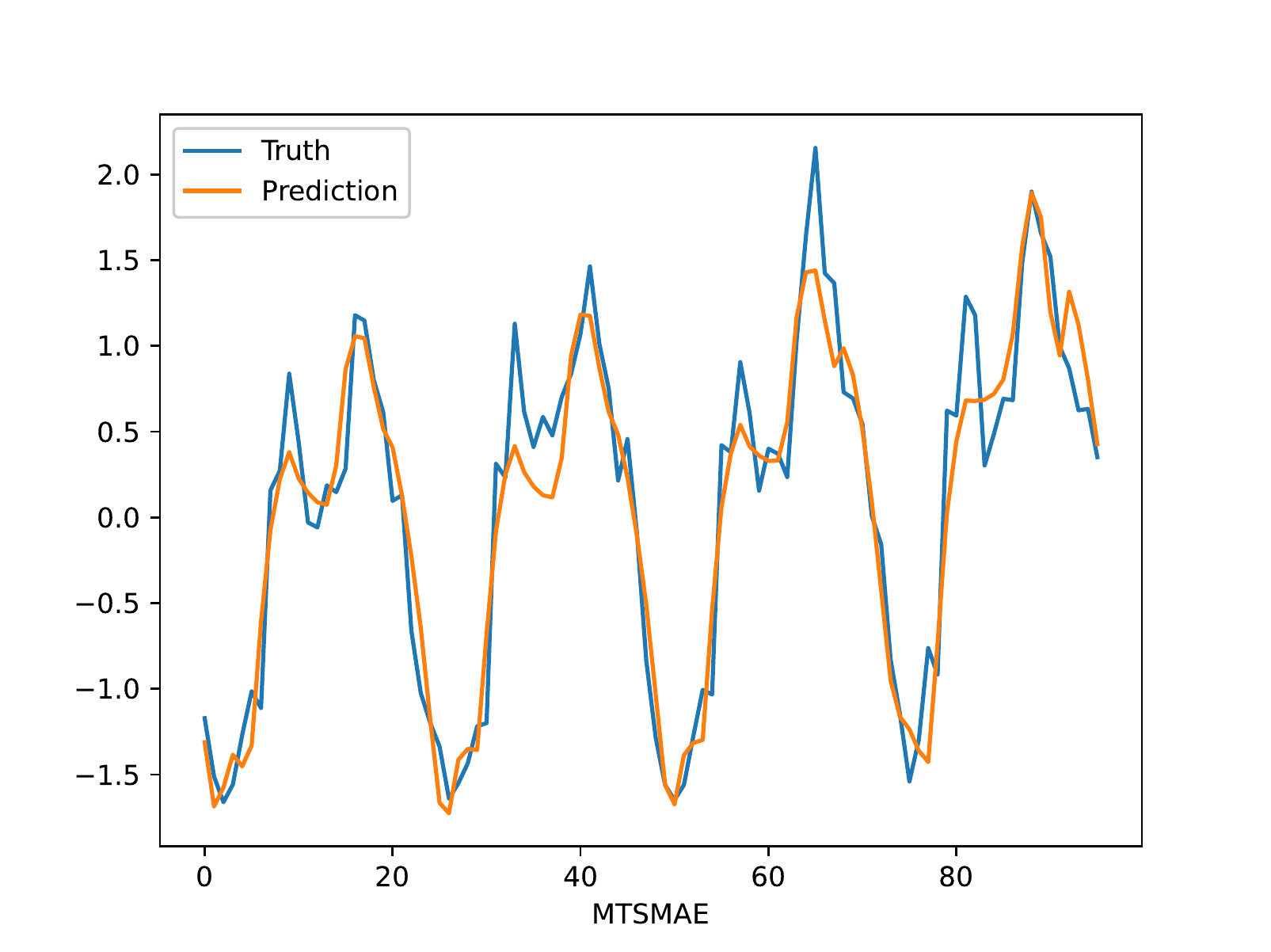}
	}
	\subfigure[Informer]{
		\includegraphics[scale=0.33]{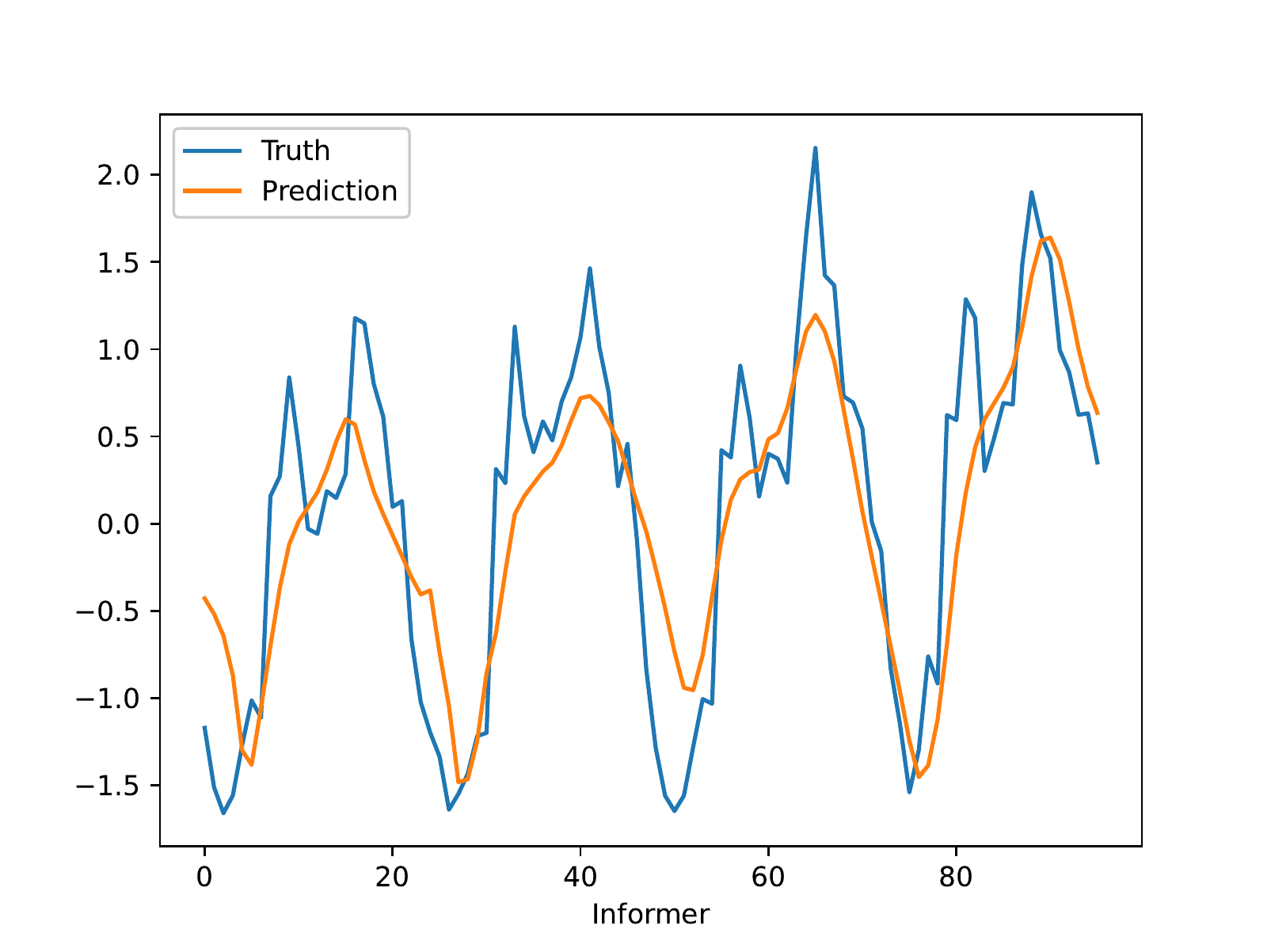}
	}
	\subfigure[LogTrans]{
		\includegraphics[scale=0.33]{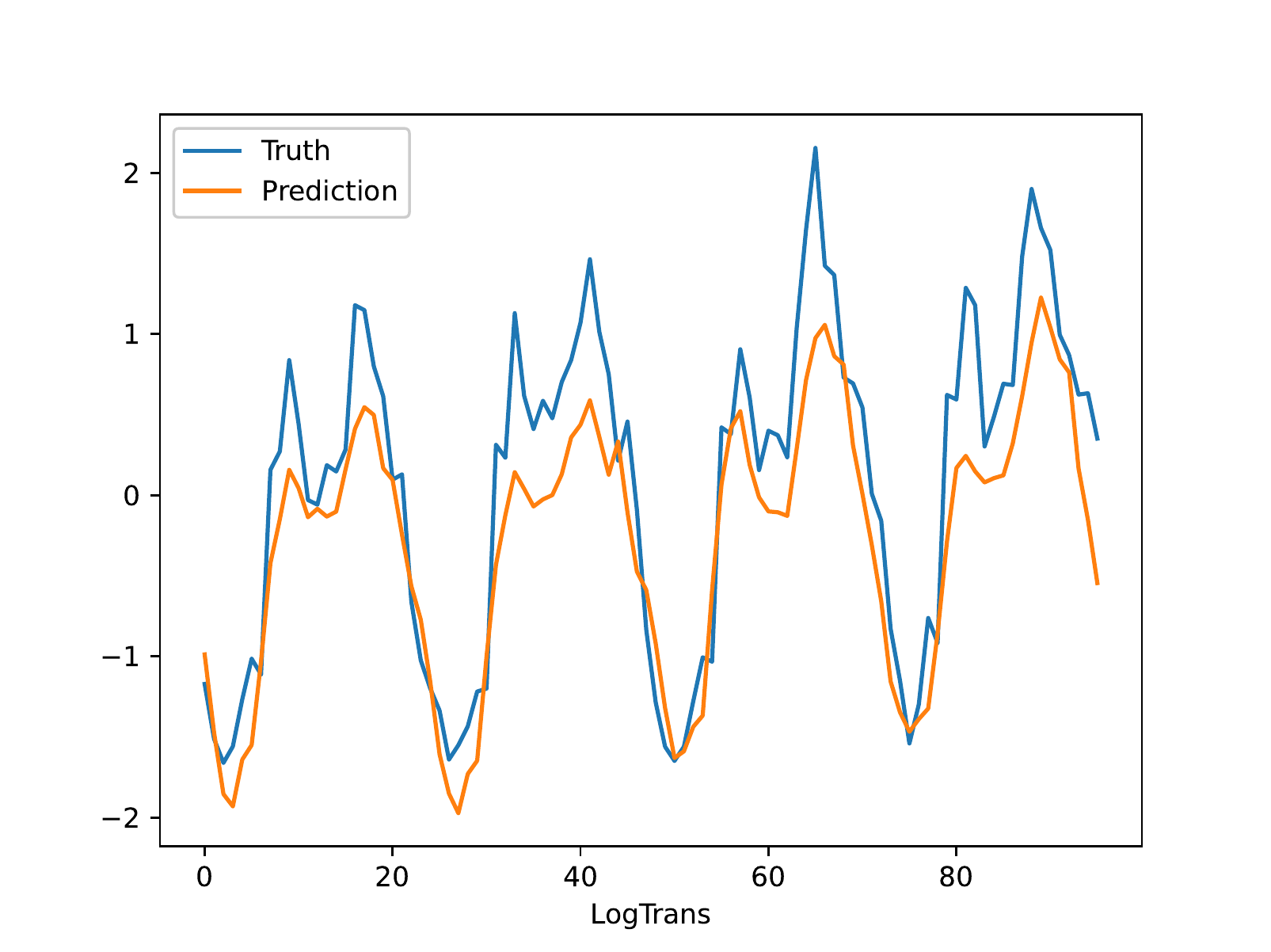}
	}
	\subfigure[LSTM]{
		\includegraphics[scale=0.33]{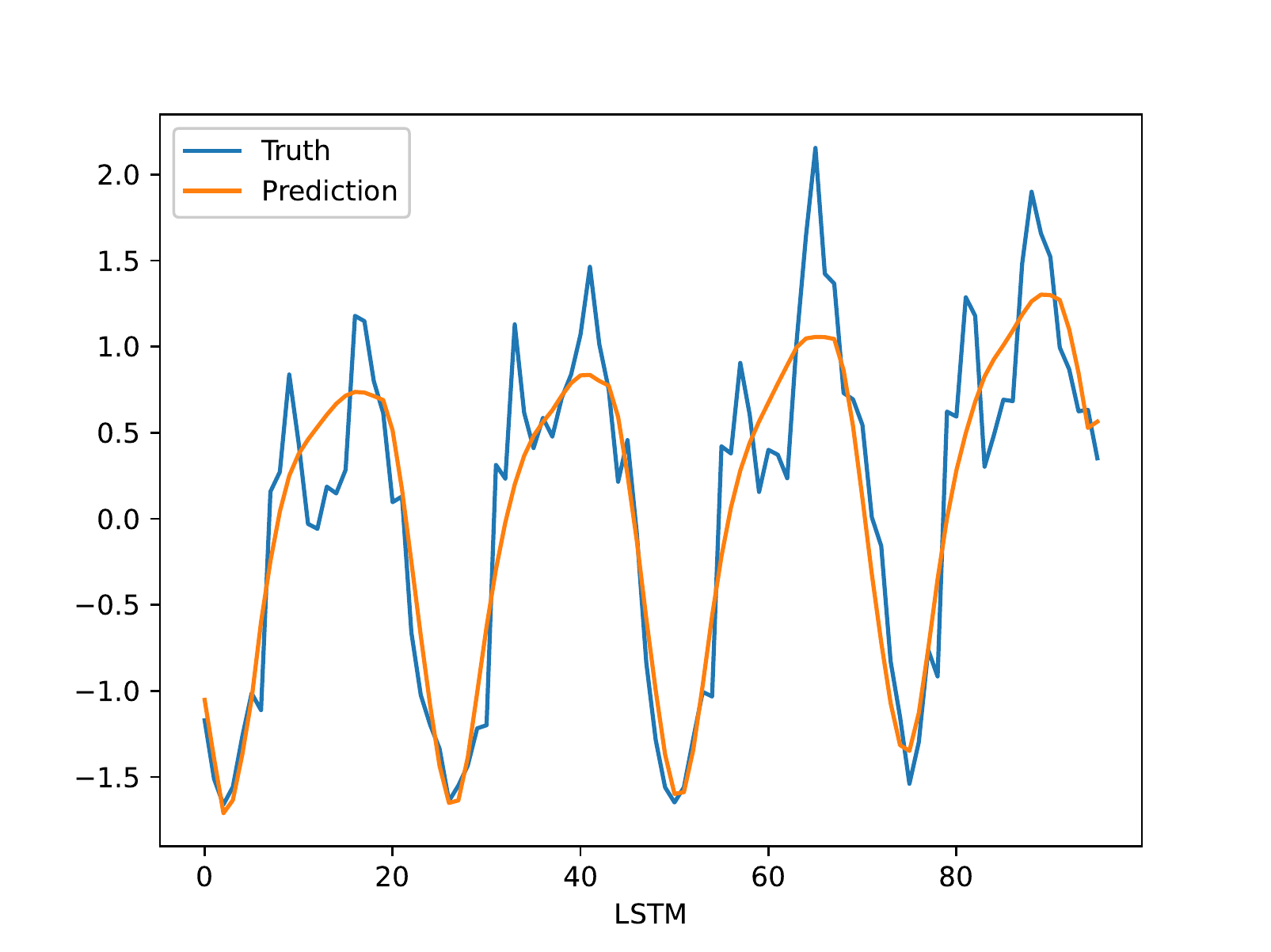}
	}
	\subfigure[Transformer]{
		\includegraphics[scale=0.33]{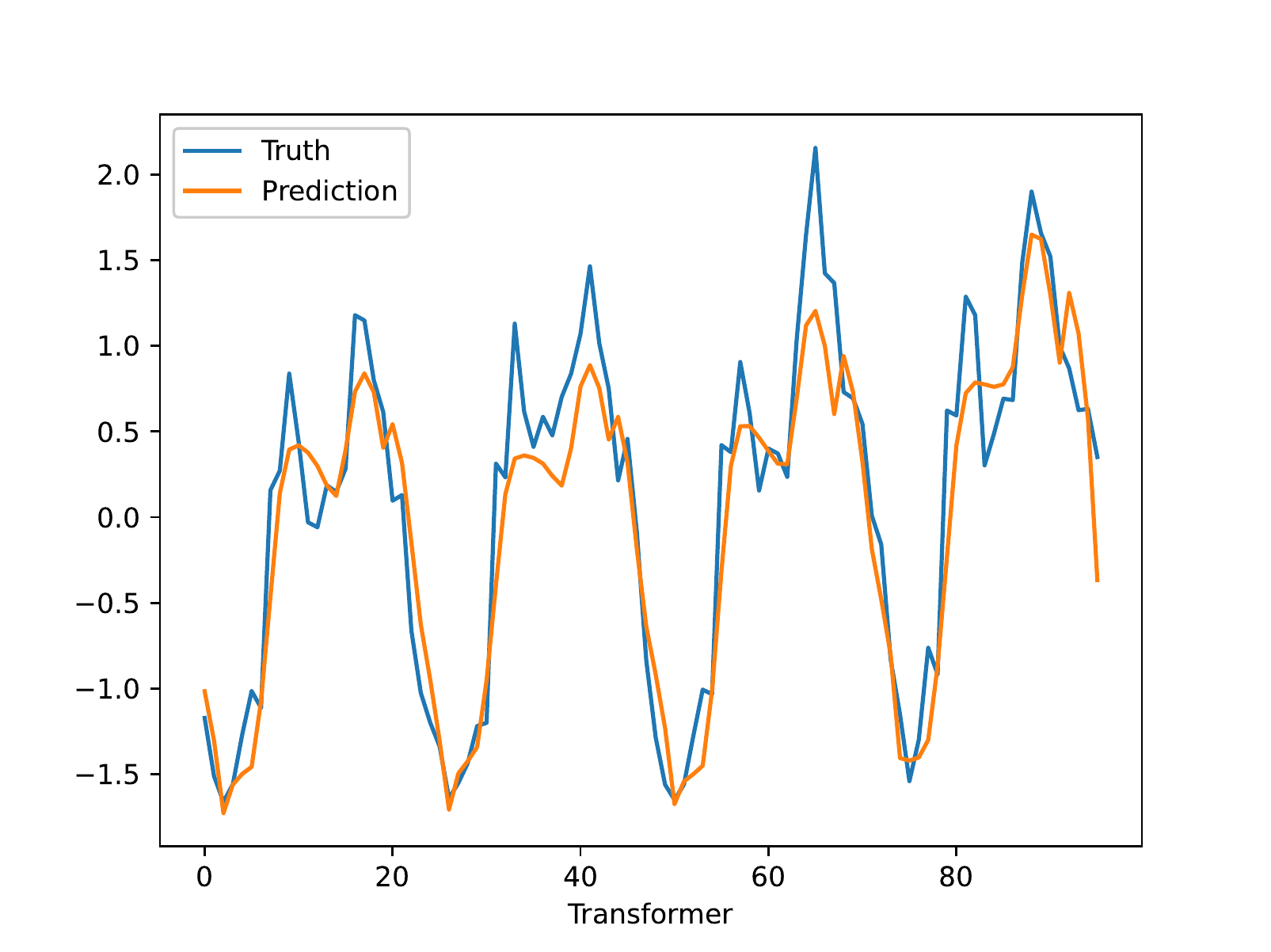}
	}
	\subfigure[Reformer]{
		\includegraphics[scale=0.33]{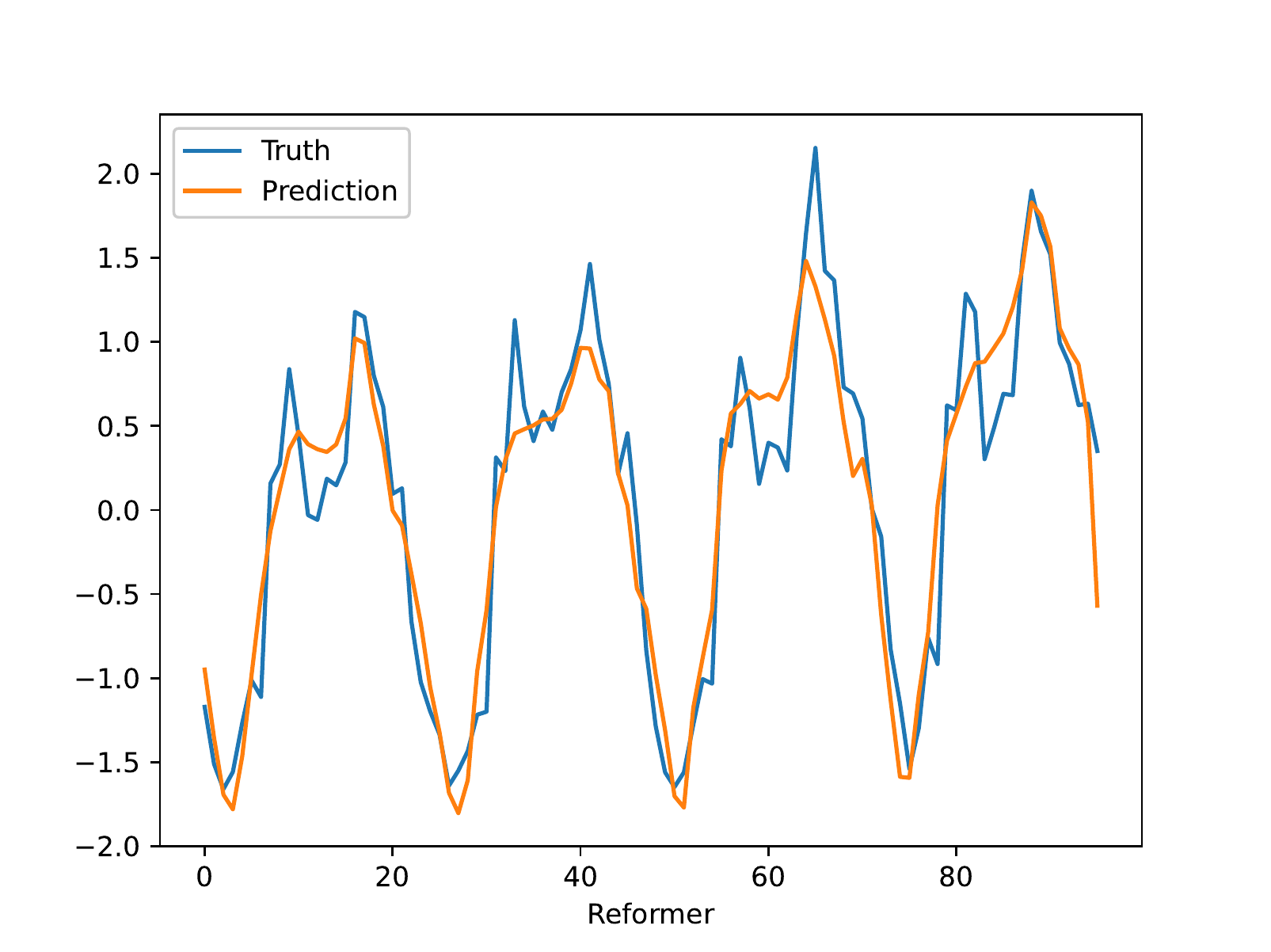}
	}
	
	\caption{Prediction cases from the ECL dataset under the $L_y=96$ setting. Blue lines are the ground truth and orange lines are the model prediction.}
	\label{fig7}
\end{figure*}

\subsubsection{Baselines}
We selected five methods as comparison, including Transformer \cite{vaswani2017attention}, three latest state-of-the-art transformer-based models: Reformer \cite{kitaev2019reformer}, LogTrans \cite{li2019enhancing}, Informer \cite{zhou2021informer}, and one RNN-based models: LSTM (Long Short-Term Memory networks) \cite{hochreiter1997long}.
\subsubsection{Experiment setting}
We fixed the random seed to ensure the repeatability of the experiment. Our experiment was implemented in Pytoch \cite{paszke2019pytorch}, and all the experiments are conducted on a single Nvidia RTX 3090 GPU (24GB memory). The input of each dataset is zero-mean normalized. We use two evaluation metrics, including mean square error (MSE) and mean absolute error (MAE):
\begin{equation}MSE= \frac{1}{n}  \sum_{i=1}^{n} \sum_{j=1}^{d} \frac{(y-\hat{y} )^2}{d} \end{equation}
 \begin{equation}MAE= \frac{1}{n}  \sum_{i=1}^{n} \sum_{j=1}^{d} \frac{\left | y-\hat{y}  \right |}{d}  \end{equation}
Where $n$ is the length of the sequence and $d$ is the dimension of data at each time point of MTSD. We use these two evaluation metrics on each prediction window to calculate the average of multivariate forecasts and roll the whole set with $stride=1$.

\begin{table}[h]
	\centering
	\renewcommand{\arraystretch}{1.2}
	\caption{Pre-training setting.}
	\label{tab0}
	\resizebox{0.4\textwidth}{!}{%
		\begin{tabular}{l|l}
			config                 & value                    \\ \hline
			optimizer              & AdamW \cite{loshchilov2017decoupled}          \\
			base learning rate     & 1e-3                   \\
			actual learning rate    & 2.5e-4                   \\
			weight decay           & 0.05                     \\
			optimizer momentum     &  $\beta_1$, $\beta_2$=0.9, 0.95 \cite{chen2020generative} \\
			batch size             & 64                       \\
			learning rate schedule & cosine decay  \cite{loshchilov2016sgdr}   \\
			loss function          & MSE    \\
			input length          &784      \\      
			patch size          &4      \\     
			encoder layers         &3      \\     
			decoder layers         &1     \\       
			masking ratio        &85\%     \\  
		\end{tabular}%
	}
\end{table}

The pre-training default setting is in Table \ref{tab0}. The input sequence length in the model is 784, and the patch size is 4, that is, we have a total of 196 patches. Due to the limitation of GPU memory, we changed the batch size to 64, and then we chose the Adam optimizer for optimization with the same strategy as MAE, that is, the configuration of the optimizer has not been modified. We follow the setting of the original MAE and use the linear learning rate scaling rule \cite{goyal2017accurate}:$lr = base\_lr \times  batchsize /256$. The number of encoder layers is 3 and the number of decoder layers is 1. In pre-training, we choose a high proportion of masking ratio by about 85\%. These configurations are the same on all datasets without making corresponding changes according to different datasets.

\begin{table}[h]
	\centering
	\renewcommand{\arraystretch}{1.2}
	\caption{Fine-tuning setting}
	\label{tab1}
	\resizebox{0.4\textwidth}{!}{%
		\begin{tabular}{l|l}
			config                 & value                    \\ \hline
			optimizer              & Adam  \cite{kingma2014adam}         \\
			initial learning rate  & 1e-4                   \\
			optimizer momentum     &  $\beta_1$, $\beta_2$=0.9, 0.999 \\
			batch size             & 32                     \\
			early stopping          & 3                     \\
			learning rate schedule & exponential decay \\
			loss function          & MSE                     
		\end{tabular}%
	}
\end{table}

\begin{table*}[th]
	
	\caption{MSTD forecasting results on four datasets}
	\renewcommand{\arraystretch}{1.2}
	\label{tab2}
	\begin{center}
		\begin{tabular}{cccccccccccccccc}
			\hline
			\multicolumn{2}{c|}{Methods}                                            & \multicolumn{2}{c|}{MTSMAE}                           & \multicolumn{2}{c|}{PatchTrans}                  & \multicolumn{2}{c|}{Informer}       & \multicolumn{2}{c|}{Reformer}                         & \multicolumn{2}{c|}{LogTrans}                         & \multicolumn{2}{c|}{Transformer}    & \multicolumn{2}{c}{LSTM} \\ \hline
			\multicolumn{2}{c|}{Metric}                                            & MSE            & \multicolumn{1}{c|}{MAE}            & MSE            & \multicolumn{1}{c|}{MAE}   & MSE   & \multicolumn{1}{c|}{MAE}   & MSE            & \multicolumn{1}{c|}{MAE}            & MSE            & \multicolumn{1}{c|}{MAE}            & MSE   & \multicolumn{1}{c|}{MAE}   & MSE    & MAE             \\ \hline
			\multicolumn{1}{c|}{\multirow{5}{*}{ECL}}   & \multicolumn{1}{c|}{24}  & 0.237          & \multicolumn{1}{c|}{0.344}          & 0.243          & \multicolumn{1}{c|}{0.347} & 0.264 & \multicolumn{1}{c|}{0.362} & 0.294          & \multicolumn{1}{c|}{0.393}          & \textbf{0.223} & \multicolumn{1}{c|}{\textbf{0.332}} & 0.242 & \multicolumn{1}{c|}{0.350} & 0.341  & 0.417           \\ 
			\multicolumn{1}{c|}{}                       & \multicolumn{1}{c|}{48}  & \textbf{0.257} & \multicolumn{1}{c|}{\textbf{0.357}} & 0.278          & \multicolumn{1}{c|}{0.369} & 0.299 & \multicolumn{1}{c|}{0.389} & 0.304          & \multicolumn{1}{c|}{0.397}          & 0.291          & \multicolumn{1}{c|}{0.374}          & 0.261 & \multicolumn{1}{c|}{0.359} & 0.335  & 0.412           \\ 
			\multicolumn{1}{c|}{}                       & \multicolumn{1}{c|}{96}  & \textbf{0.263} & \multicolumn{1}{c|}{\textbf{0.362}} & 0.272          & \multicolumn{1}{c|}{0.370} & 0.302 & \multicolumn{1}{c|}{0.388} & 0.293          & \multicolumn{1}{c|}{0.382}          & 0.299          & \multicolumn{1}{c|}{0.379}          & 0.286 & \multicolumn{1}{c|}{0.380} & 0.329  & 0.408           \\ 
			\multicolumn{1}{c|}{}                       & \multicolumn{1}{c|}{192} & \textbf{0.278} & \multicolumn{1}{c|}{0.371}          & 0.328          & \multicolumn{1}{c|}{0.415} & 0.322 & \multicolumn{1}{c|}{0.379} & 0.349          & \multicolumn{1}{c|}{0.426}          & 0.279          & \multicolumn{1}{c|}{\textbf{0.370}} & \multicolumn{2}{c|}{-}             & 0.326  & 0.407           \\ 
			\multicolumn{1}{c|}{}                       & \multicolumn{1}{c|}{384} & \textbf{0.289} & \multicolumn{1}{c|}{\textbf{0.382}} & 0.368          & \multicolumn{1}{c|}{0.441} & 0.338 & \multicolumn{1}{c|}{0.411} & \multicolumn{2}{c|}{-}                               & 0.331          & \multicolumn{1}{c|}{0.404}          & \multicolumn{2}{c|}{-}             & 0.319  & 0.402           \\ \hline
			\multicolumn{1}{c|}{\multirow{5}{*}{ETTm1}} & \multicolumn{1}{c|}{24}  & \textbf{0.282} & \multicolumn{1}{c|}{\textbf{0.356}} & 0.311          & \multicolumn{1}{c|}{0.379} & 0.362 & \multicolumn{1}{c|}{0.407} & 0.397          & \multicolumn{1}{c|}{0.423}          & 0.905          & \multicolumn{1}{c|}{0.656}          & 0.339 & \multicolumn{1}{c|}{0.392} & 1.253  & 0.837           \\ 
			\multicolumn{1}{c|}{}                       & \multicolumn{1}{c|}{48}  & \textbf{0.412} & \multicolumn{1}{c|}{\textbf{0.445}} & 0.455          & \multicolumn{1}{c|}{0.477} & 0.461 & \multicolumn{1}{c|}{0.458} & 0.558          & \multicolumn{1}{c|}{0.535}          & 0.880          & \multicolumn{1}{c|}{0.662}          & 0.582 & \multicolumn{1}{c|}{0.579} & 1.262  & 0.838           \\ 
			\multicolumn{1}{c|}{}                       & \multicolumn{1}{c|}{96}  & \textbf{0.558} & \multicolumn{1}{c|}{\textbf{0.544}} & 0.657          & \multicolumn{1}{c|}{0.604} & 0.618 & \multicolumn{1}{c|}{0.576} & 0.703          & \multicolumn{1}{c|}{0.611}          & 0.562          & \multicolumn{1}{c|}{0.553}          & 0.855 & \multicolumn{1}{c|}{0.752} & 1.197  & 0.829           \\ 
			\multicolumn{1}{c|}{}                       & \multicolumn{1}{c|}{192} & 0.603          & \multicolumn{1}{c|}{0.597}          & \textbf{0.531} & \multicolumn{1}{c|}{0.546} & 0.765 & \multicolumn{1}{c|}{0.659} & 0.867          & \multicolumn{1}{c|}{0.678}          & 0.564          & \multicolumn{1}{c|}{\textbf{0.527}} & \multicolumn{2}{c|}{-}             & 1.200  & 0.824           \\
			\multicolumn{1}{c|}{}                       & \multicolumn{1}{c|}{384} & \textbf{0.672} & \multicolumn{1}{c|}{0.634}          & 0.818          & \multicolumn{1}{c|}{0.708} & 0.965 & \multicolumn{1}{c|}{0.763} & \multicolumn{2}{c|}{-}                               & 0.694          & \multicolumn{1}{c|}{\textbf{0.612}} & \multicolumn{2}{c|}{-}             & 1.199  & 0.822           \\ \hline
			\multicolumn{1}{c|}{\multirow{5}{*}{ETTh2}} & \multicolumn{1}{c|}{24}  & \textbf{0.648} & \multicolumn{1}{c|}{\textbf{0.661}} & 0.689          & \multicolumn{1}{c|}{0.686} & 1.567 & \multicolumn{1}{c|}{1.022} & 0.893          & \multicolumn{1}{c|}{0.764}          & 0.851          & \multicolumn{1}{c|}{0.725}          & 0.968 & \multicolumn{1}{c|}{0.802} & 3.239  & 1.371           \\ 
			\multicolumn{1}{c|}{}                       & \multicolumn{1}{c|}{48}  & \textbf{1.126} & \multicolumn{1}{c|}{\textbf{0.874}} & 1.266          & \multicolumn{1}{c|}{0.948} & 3.326 & \multicolumn{1}{c|}{1.500} & 1.240          & \multicolumn{1}{c|}{0.940}          & 1.165          & \multicolumn{1}{c|}{0.904}          & 2.123 & \multicolumn{1}{c|}{1.218} & 3.359  & 1.403           \\
			\multicolumn{1}{c|}{}                       & \multicolumn{1}{c|}{96}  & \textbf{2.561} & \multicolumn{1}{c|}{\textbf{1.304}} & 3.167          & \multicolumn{1}{c|}{1.555} & 7.756 & \multicolumn{1}{c|}{2.321} & 4.161          & \multicolumn{1}{c|}{1.637}          & 2.795          & \multicolumn{1}{c|}{1.431}          & 6.597 & \multicolumn{1}{c|}{2.163} & 3.309  & 1.391           \\ 
			\multicolumn{1}{c|}{}                       & \multicolumn{1}{c|}{192} & \textbf{3.341} & \multicolumn{1}{c|}{1.588}          & 4.687          & \multicolumn{1}{c|}{1.800} & 5.527 & \multicolumn{1}{c|}{1.993} & 4.425          & \multicolumn{1}{c|}{1.791}          & 5.135          & \multicolumn{1}{c|}{1.946}          & \multicolumn{2}{c|}{-}             & 3.417  & \textbf{1.415}  \\ 
			\multicolumn{1}{c|}{}                       & \multicolumn{1}{c|}{384} & \textbf{3.107} & \multicolumn{1}{c|}{\textbf{1.403}} & 3.375          & \multicolumn{1}{c|}{1.626} & 3.581 & \multicolumn{1}{c|}{1.587} & \multicolumn{2}{c|}{-}                               & 3.156          & \multicolumn{1}{c|}{1.561}          & \multicolumn{2}{c|}{-}             & 3.456  & 1.422           \\ \hline
			\multicolumn{1}{c|}{\multirow{5}{*}{WTH}}   & \multicolumn{1}{c|}{24}  & \textbf{0.301} & \multicolumn{1}{c|}{\textbf{0.360}} & 0.305          & \multicolumn{1}{c|}{0.367} & 0.366 & \multicolumn{1}{c|}{0.421} & 0.306          & \multicolumn{1}{c|}{0.360}          & 0.387          & \multicolumn{1}{c|}{0.436}          & 0.348 & \multicolumn{1}{c|}{0.390} & 0.941  & 0.771           \\ 
			\multicolumn{1}{c|}{}                       & \multicolumn{1}{c|}{48}  & \textbf{0.377} & \multicolumn{1}{c|}{\textbf{0.414}} & 0.389          & \multicolumn{1}{c|}{0.423} & 0.478 & \multicolumn{1}{c|}{0.490} & 0.391          & \multicolumn{1}{c|}{0.428}          & 0.501          & \multicolumn{1}{c|}{0.519}          & 0.449 & \multicolumn{1}{c|}{0.459} & 0.946  & 0.773           \\ 
			\multicolumn{1}{c|}{}                       & \multicolumn{1}{c|}{96}  & 0.496          & \multicolumn{1}{c|}{0.501}          & 0.513          & \multicolumn{1}{c|}{0.514} & 0.613 & \multicolumn{1}{c|}{0.583} & \textbf{0.469} & \multicolumn{1}{c|}{\textbf{0.480}} & 0.551          & \multicolumn{1}{c|}{0.541}          & 0.536 & \multicolumn{1}{c|}{0.517} & 0.943  & 0.772           \\ 
			\multicolumn{1}{c|}{}                       & \multicolumn{1}{c|}{192} & \textbf{0.546} & \multicolumn{1}{c|}{\textbf{0.534}} & 0.589          & \multicolumn{1}{c|}{0.557} & 0.665 & \multicolumn{1}{c|}{0.607} & 0.564          & \multicolumn{1}{c|}{0.531}          & 0.577          & \multicolumn{1}{c|}{0.558}          & \multicolumn{2}{c|}{-}             & 0.945  & 0.772           \\ 
			\multicolumn{1}{c|}{}                       & \multicolumn{1}{c|}{384} & \textbf{0.573} & \multicolumn{1}{c|}{\textbf{0.563}} & 0.577          & \multicolumn{1}{c|}{0.564} & 0.660 & \multicolumn{1}{c|}{0.612} & \multicolumn{2}{c|}{-}                               & 0.635          & \multicolumn{1}{c|}{0.591}          & \multicolumn{2}{c|}{-}             & 0.939  & 0.769           \\ \hline
			\multicolumn{2}{c|}{Count}                                             & \multicolumn{2}{c|}{31}                              & \multicolumn{2}{c|}{1}                      & \multicolumn{2}{c|}{0}             & \multicolumn{2}{c|}{2}                               & \multicolumn{2}{c|}{5}                               & \multicolumn{2}{c|}{0}             & \multicolumn{2}{c}{1}   \\ \hline
		\end{tabular}
		
	\end{center}
\end{table*}

Our fine-tuning follows common practice of Informer training, and the default setting is in Table \ref{tab1}. In the fine-tuning, we use Adam optimizer for optimization with a learning rate starts from $1e^{-4}$, decaying two times smaller every epoch. There is no limit to the total number of epochs, with appropriate early stopping, that is, when the loss of the validation set does not decrease on the three epochs, the training will be stopped. We set the comparison methods according to the suggestion, and the batch size is 32.

\subsection{Results and Analysis}

\subsubsection{Multivariate Time-series Forecasting}

In order to be consistent with the setting of pre-training, we choose the same input length, i.e. $L_x=784$, and we gradually extend the size of the prediction window $L_y$ as a higher requirement of prediction capacity, i.e., \{24, 48, 96, 192, 384\}, representing \{6h, 12h, 24h, 48h, 96h\} in ETTm1, \{1d, 2d, 4d, 8d, 16d\} in \{ETTh2, ECL, WTH\}, and we set the length of the label to double $L_y$, i.e., \{48, 96, 192, 384, 768\}. The parameter sensitivity of input length and output length will be discussed in subsequent chapters. In the pre-training, the default configuration selected by our MTSMAE is shown in the Table \ref{tab0},

Table \ref{tab2} summarizes the multivariable evaluation results of all metrics on \{ECL, ETTm1, ETTh2, WTH\}. The best results are highlighted in bold, where `-' represents out-of-memory, and PatchTrans indicates our model without pre-training.
From Table \ref{tab2}, we observe that the proposed model MTSMAE has better performance than other methods. Compared with PatchTrans without pre-training, the MTSMAE improves the prediction performance in almost all datasets with different prediction lengths (except in the case of $L_y=192$ in the ETTm1 dataset). 

In order to better evaluate the prediction ability of different models, we drew one dimension of the prediction results of the $test$ $set$ from the ECL dataset for qualitative comparison. Fig.~\ref{fig7} intuitively shows the prediction results of different models. All models can roughly predict the development trend of MTSD, but our model shows the best performance in different models. Since prediction results are more consistent with the ground truth, our model can accurately predict the periodicity and long-term changes. Compared with other Transformer based models, prediction results of LSTM are smoother, so it can’t well predict the trend of some inflection points, which is why LSTM can’t be used in long time-series prediction.

\begin{table*}[th]
	\centering
	\caption{Different input lengths for two prediction lengths}
	\label{tab4}
	\renewcommand{\arraystretch}{1.3}
	\begin{tabular}{cc|cccccc|ccccc} \hline
		\multicolumn{2}{c|}{Predicition length} & \multicolumn{6}{c|}{336}                                                                                         & \multicolumn{5}{c}{480}                                                                            \\\hline
		\multicolumn{2}{c|}{Encoder's input}    & 336            & 480            & 720            & 960                & 1200               & 1440               & 480            & 720                & 960                & 1200               & 1440               \\\hline
		\multirow{2}{*}{Informer}     & MSE    & 3.339          & 3.256          & 3.423          & 3.460              & 3.371              & 3.099              & 3.374          & 3.504              & 3.441              & 3.441              & 3.400              \\
		& MAE    & 1.393          & 1.379          & \textbf{1.421} & 1.418              & 1.401              & \textbf{1.337}     & 1.403          & 1.434              & 1.421              & 1.417              & 1.408              \\\hline
		\multirow{2}{*}{Reformer}     & MSE    & \textbf{2.471} & 3.692          & 4.561          & 3.479              & 3.677              & \multirow{2}{*}{-} & 3.078          & 3.052              & 2.873              & 3.271              & \multirow{2}{*}{-} \\
		& MAE    & \textbf{1.229} & \textbf{1.482} & 1.692          & 1.489              & 1.664              &                    & 1.381          & 1.473              & 1.451              & 1.324              &                    \\\hline
		\multirow{2}{*}{LogTrans}     & MSE    & 3.217          & 3.917          & 3.391          & \multirow{2}{*}{-} & \multirow{2}{*}{-} & \multirow{2}{*}{-} & 3.431          & \multirow{2}{*}{-} & \multirow{2}{*}{-} & \multirow{2}{*}{-} & \multirow{2}{*}{-} \\
		& MAE    & 1.461          & 1.655          & 1.554          &                    &                    &                    & 1.579          &                    &                    &                    &                    \\\hline
		\multirow{2}{*}{PatchTrans}       & MSE    & 3.588          & 3.996          & 3.919          & 2.963              & 2.937              & 3.588              & 3.453          & 3.069              & 2.900              & 2.789              & 3.087              \\
		& MAE    & 1.613          & 1.681          & 1.673          & 1.465              & 1.408              & 1.563              & 1.615          & 1.423              & 1.438              & 1.370              & 1.451              \\\hline
		\multirow{2}{*}{MTSMAE}       & MSE    & 3.507          & \textbf{3.025} & \textbf{3.386} & \textbf{2.946}     & \textbf{2.691}     & \textbf{2.883}     & \textbf{2.974} & \textbf{2.692}     & \textbf{2.769}     & \textbf{2.746}     & \textbf{2.575}     \\
		& MAE    & 1.572          & 1.492          & 1.569          & \textbf{1.397}     & \textbf{1.377}     & 1.365              & 1.506          & \textbf{1.373}     & \textbf{1.365}     & \textbf{1.313}     & \textbf{1.362}    \\\hline
	\end{tabular}%

\end{table*}

\subsubsection{Input Length}

We perform the sensitivity analysis (input length) of the proposed MTSMAE model on ETTh2 dataset. Compared with the previous experiment, $L_x=784$, in this experiment, we gradually extended the length of the input sequence, i.e., \{336, 480, 720, 960, 1200, 1440\}, while keeping the prediction sequence unchanged. We choose two prediction lengths, 336 and 480, and the length of the label sequence is consistent with the length of the prediction sequence.

The experiment results are shown in the Table ~\ref{tab4}. It can be shown that increasing the input length will lead to the decline of MSE and MAE, because it brings repeated short-term patterns, and longer encoder inputs may contain more dependencies. However, with the increasing of input sequence, although the input information will increase, the influence of noise will also increase. Some models can not effectively remove the impact of these noises, so MSE and MAE may increase. The performance of our proposed MTSMAE in short sequence input is not the best, but with the increase of input sequence, the prediction performance of MTSMAE is getting better and better.
There is no problem of out-of-memory due to the long input sequence and sudden decline in prediction performance, which demonstrates that MTSMAE has good denoising ability.

\subsubsection{Masking ratio}

\begin{figure*}[th]
	\centering
	\subfigure[MSE]{
		\includegraphics[scale=0.52]{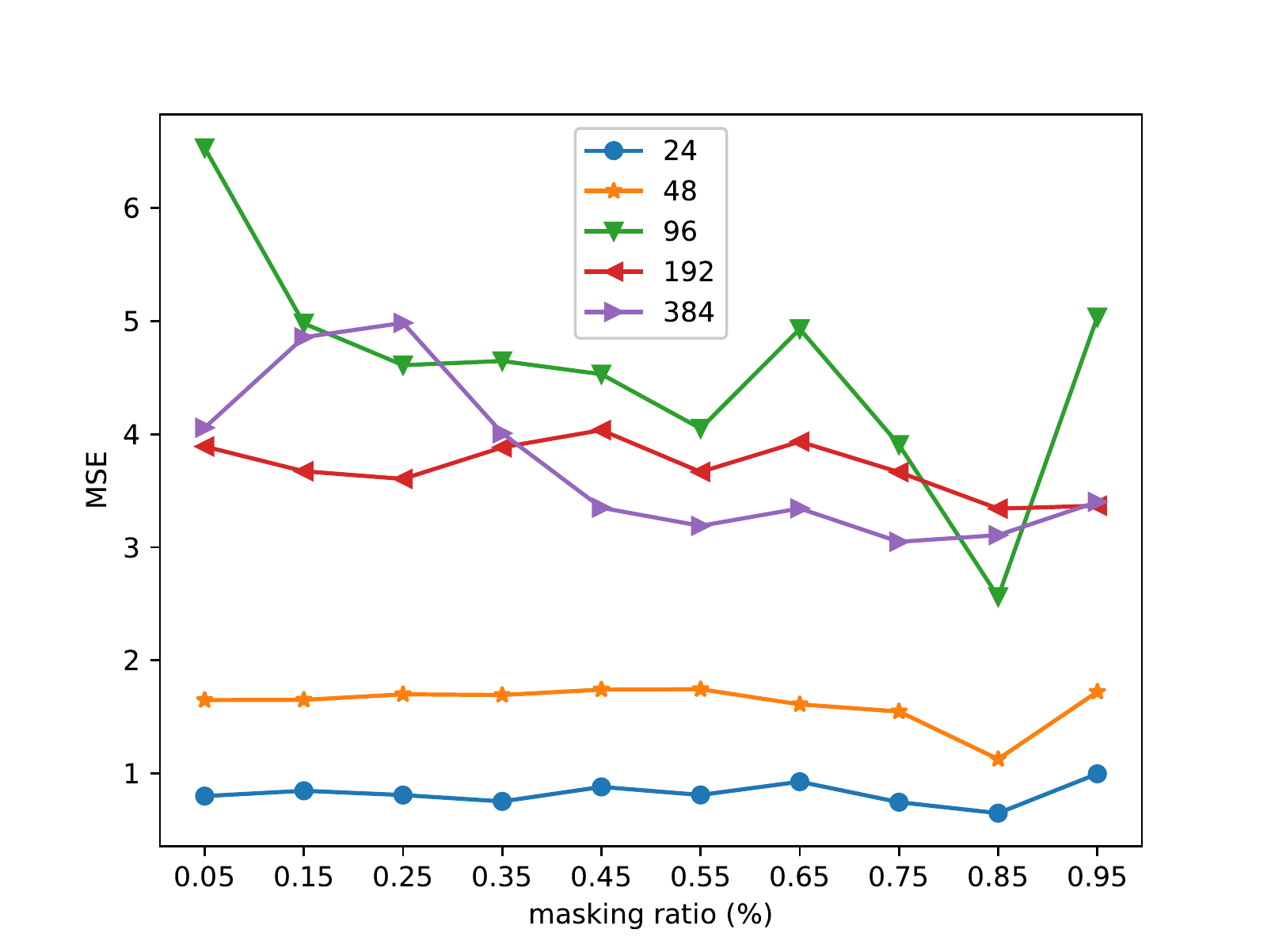}
		\label{fig4}
	}
	\subfigure[MAE]{
		\includegraphics[scale=0.52]{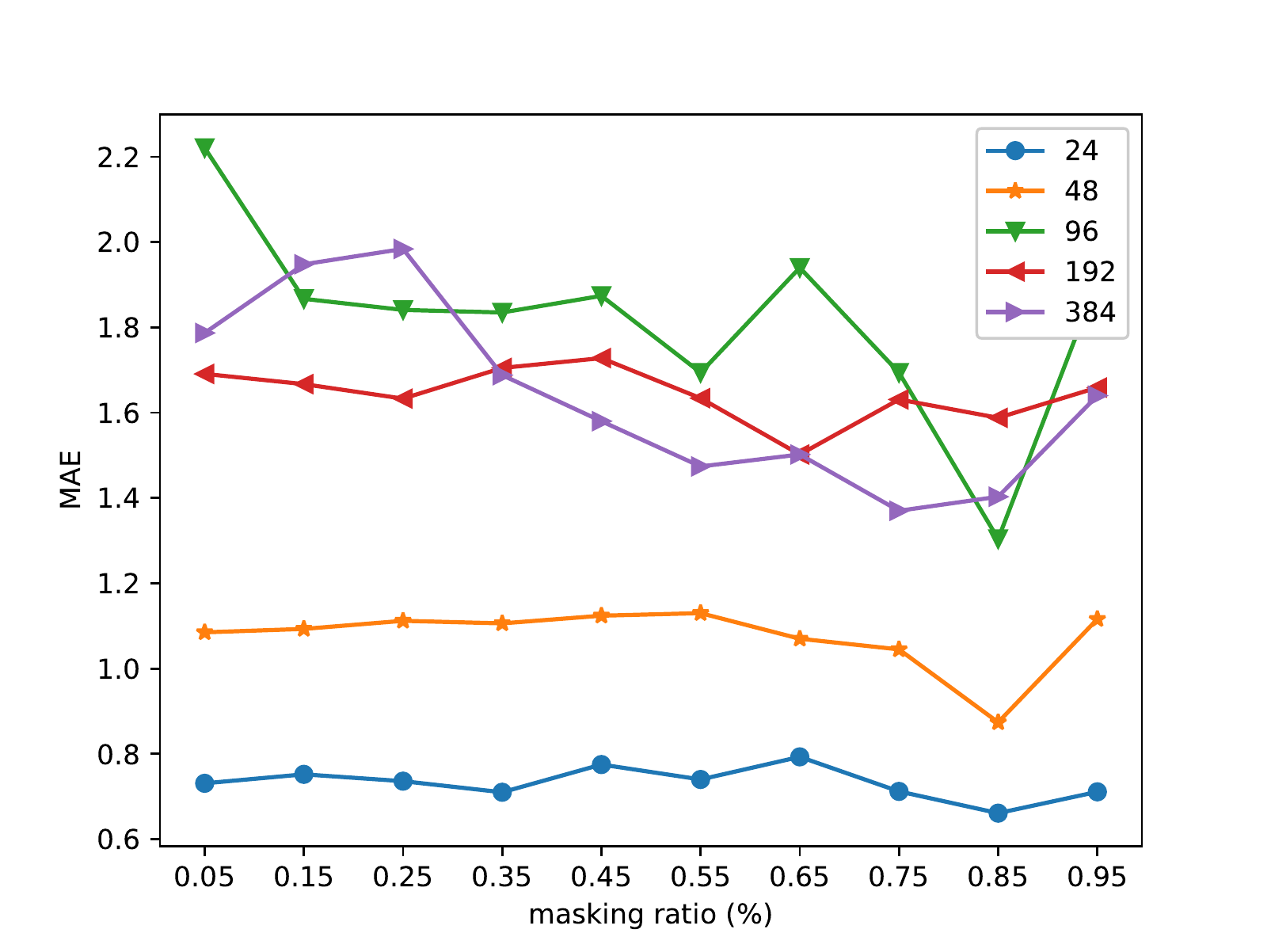}
		\label{fig5}
	}
	\caption{Results of MSE and MAE at different prediction lengths with different masking ratios.}
	\label{fig6}
\end{figure*}

Fig.~\ref{fig6} shows the influence of different masking ratio. The dataset of this experiment is ETTh2. The default encoder and decoder layers are selected during pre-training, i.e., the number of decoder layers is 3 and decoder layers is 1. Different broken lines in the figure represent different prediction lengths. The optimal ratio is beyond our expectation, which is higher than the original MAE's optimal ratios (75\%), and this masking ratio is also opposite to the self-supervised method BERT in the NLP field, which selects a masking ratio of 15\% (a very low masking ratio). 

High ratio masking is conducive to the fine-tuning of subsequent downstream tasks, which related to the data characteristics of multivariate time-series. High ratio masking can lose more information, greatly reduce the redundancy of data, and increase the overall understanding of the model beyond low-level information. However, a too high masking ratio, such as 95\%, means that a large amount of data is lost. The data that the model can learn is finite, which is not conducive to the model's understanding of the data.

\subsubsection{Decoder depth}

\begin{table}[]	
	\caption{Different decoder depths}
	\begin{center}
		\renewcommand{\arraystretch}{1.2}
		\resizebox{0.4\textwidth}{!}{%
			\begin{tabular}{cc|c|c|c|c}\hline
				\multicolumn{2}{c|}{Blocks}                     & 1     & 2     & 3     & 4     \\\hline
				\multicolumn{1}{c|}{\multirow{2}{*}{24}}  & MSE & 0.648 & 0.619 & 0.609 & 0.584 \\ 
				\multicolumn{1}{c|}{}                     & MAE & 0.661 & 0.644 & 0.649 & 0.628 \\\hline
				\multicolumn{1}{c|}{\multirow{2}{*}{48}}  & MAE & 1.126 & 1.058 & 0.975 & 1.111 \\
				\multicolumn{1}{c|}{}                     & MAE & 0.874 & 0.824 & 0.807 & 0.856 \\\hline
				\multicolumn{1}{c|}{\multirow{2}{*}{96}}  & MSE & 2.561 & 2.680 & 2.461 & 2.806 \\
				\multicolumn{1}{c|}{}                     & MAE & 1.304 & 1.380 & 1.306 & 1.346 \\\hline
				\multicolumn{1}{c|}{\multirow{2}{*}{192}} & MSE & 3.341 & 3.262 & 3.501 & 3.752 \\
				\multicolumn{1}{c|}{}                     & MAE & 1.588 & 1.571 & 1.541 & 1.566 \\\hline
				\multicolumn{1}{c|}{\multirow{2}{*}{384}} & MSE & 3.107 & 2.931 & 2.787 & 2.932 \\
				\multicolumn{1}{c|}{}                     & MAE & 1.403 & 1.933 & 1.344 & 1.398\\\hline
			\end{tabular}
		}
		\label{tab3}
	\end{center}
\end{table}

As shown in the Table ~\ref{tab3}, our MTSMAE decoder can be designed flexibly. We changed the depth of the decoder (the number of Transformer decoder blocks), i.e., \{1, 2, 3, 4 \}, keep the fixed depth of the encoder as 3. The last few layers of the autoencoder are dedicated to reconstruction and have little correlation with acquiring semantics. During pre-training, our encoder is also used to recover the original data. A reasonable depth decoder can explain the specialization of reconstruction, leaving the latent representation at a more abstract level.

The dataset selected in this experiment is ETTh2, and we choose 85\% masking ratio. As shown in Table ~\ref{tab3}, the deeper the decoder has the correspondingly better performance of downstream prediction tasks. However, when the number of layers of the decoder reaches 4 (more than the number of layers of the encoder), the prediction performance of MTSMAE begins to decline. A deep enough decoder is extremely important for the downstream prediction task, but too deep decoder is easy to affect the performance of the encoder to obtain semantics.

\section{Conclusion}
In this paper, we studied the problem of multivariable time-series forecasting, and proposed a model named MTSMAE for multivariable time-series data. Specifically, firstly according to the data characteristics of multivariable time-series, we proposed a new patch embedding method, which can reduce the use of memory, so that the model can process longer sequences. Then, we further proposed a self-supervised pre-training method based on masked autoencoders (MAE).
A large number of experiments on real datasets have proved the effectiveness of MTSMAE in improving multivariable time-series forecasting.




\end{document}